\documentclass[letterpaper, 10 pt, conference]{ieeeconf}  % Comment this line out if you need a4paper

\usepackage{graphicx} % for pdf, bitmapped graphics files
\usepackage{subfigure}
\usepackage{multirow}
\usepackage{amssymb}
\usepackage{url} % assumes new font selection scheme installed
\usepackage{amsmath} % assumes amsmath package installed
\usepackage{caption}

\title{\bf Rhythmic Representations: Learning Periodic Patterns for Scalable Place Recognition at a Sub-Linear Storage Cost}

\author{Litao Yu, Adam Jacobson and Michael Milford \thanks{Litao Yu (litao.yu@qut.edu.au), Adam Jacobson (adam.jacobson@hdr.qut.edu.au) and Michael Milford (michael.milford@qut.edu.au) are with the School of Electrical Engineering and Computer Science, Queensland University of Technology, Brisbane, QLD, Australia. MM also with the Australian Centre for Robotic Vision.}
}

\begin{document}

\maketitle
%\thispagestyle{empty}
%\pagestyle{empty}

%%%%%%%%%%%%%%%%%%%%%%%%%%%%%%%%%%%%%%%%%%%%%%%%%%%%%%%%%%%%%%%%%%%%%%%%%%%%%%%%
\begin{abstract}

Robotic and animal mapping systems share many challenges and characteristics: they must function in a wide variety of environmental conditions, enable the robot or animal to navigate effectively to find food or shelter, and be computationally tractable from both a speed and storage perspective. With regards to map storage, the mammalian brain appears to take a diametrically opposed approach to all current robotic mapping systems. Where robotic mapping systems attempt to solve the data association problem to minimise representational aliasing, neurons in the brain intentionally break data association by encoding large (potentially unlimited) numbers of places with a single neuron. In this paper, we propose a novel method based on supervised learning techniques that seeks out regularly repeating visual patterns in the environment with mutually complementary co-prime frequencies, and an encoding scheme that enables storage requirements to grow sub-linearly with the size of the environment being mapped. To improve robustness in challenging real-world environments while maintaining storage growth sub-linearity, we incorporate both multi-exemplar learning and data augmentation techniques. Using large benchmark robotic mapping datasets, we demonstrate the combined system achieving high-performance place recognition with sub-linear storage requirements, and characterize the performance-storage growth trade-off curve. The work serves as the first robotic mapping system with sub-linear storage scaling properties, as well as the first large-scale demonstration in real-world environments of one of the proposed memory benefits of these neurons.

\end{abstract}

%%%%%%%%%%%%%%%%%%%%%%%%%%%%%%%%%%%%%%%%%%%%%%%%%%%%%%%%%%%%%%%%%%%%%%%%%%%%%%%%
\section{Introduction}

Visual place recognition - recognising whether a current camera image matches to those stored in a map or database - is a fundamental component of most robotic  mapping and navigation systems\cite{TR:SLAM_SURVEY}. These mapping systems are typically developed and evaluated based on the quality of the map they can produce, the robustness of representations and their associated computational requirements. Much emphasis has been placed on solving the ``data association'' problem - making sure that there are no incorrectly or aliased map-landmark associations.

\begin{figure}[t!]
\centering
\includegraphics[width=0.3\textwidth]{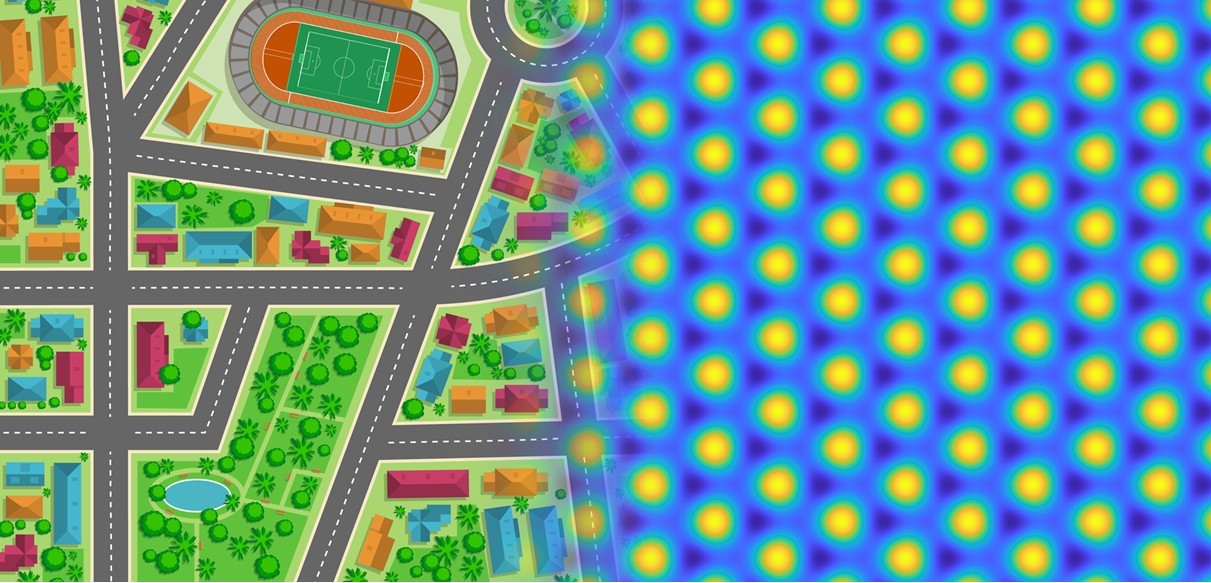}
\caption{Neurons  in the mammalian brain known as grid cells intentionally alias their representation of the environment; each neuron represents an arbitrary number of places in a regularly repeating pattern. }
\label{fig:intuation}
\end{figure}

Navigation neurons found in the brain of many mammals such as rodents, known as ``grid cells'' \cite{SCIENCE:TF} (see Fig. \ref{fig:intuation}), have highly aliased data associations with locations in the environment - each cell encodes an arbitrary number of physical locations laid out in a triangular tesselating grid \cite{NATURE:M, ARN:PGB}. There has been much interest in the theoretical advantages of such a neural representation including implications for memory storage, error correction \cite{NATUREN:GC}  and scalability that could revolutionize how artificial systems including robots are developed. 

In this paper, we propose a novel approach to discover regularly repeating visual patterns in an environment, and to encode these regularly repeated patterns in frame sequences (see Fig. \ref{fig:intuation}). We adopt a supervised learning approach and take advantage of statistical properties to identify periodicity in the world being mapped. To perform place recognition, a global location estimate is reconstructed from identifying the phase of these learned patterns in a current camera image. In this way, the storage requirements scale up sub-linearly with the number of encoded places in the environment (or better). Extensive experiments on three real-world datasets demonstrate successful place recognition while retaining sub-linear storage growth.

We present new research that significantly extends a pilot study \cite{IROS17:DEJAVU} by developing a number of new contributions that address scalability to large, challenging real world environments, including:

\begin{itemize}
  \item A method for actively finding and learning periodic visual patterns from frame sequences using supervised learning techniques and best practice for maximizing their utility in sub-linear mapping,
  \item Techniques for optimizing for minimum storage requirements that balance the number of periodic patterns, their period lengths and their separability,
  \item Development of a multi-exemplar training scheme that improves place recognition performance in perceptually challenging environments where multiple training examples are available, while maintaining sub-linear storage growth,
  \item Visual data augmentation techniques for improving performance when multiple-examples are not available, and
  \item Comprehensive performance evaluation on several large benchmark datasets including characterizations of the tradeoff between storage scalability and place recognition performance, and analysis of the benefits of multi-exemplar and augmentation-based training.
\end{itemize}

Together these contributions represent a significant step towards enabling a sub-linear, highly scalable map encoding scheme for autonomous systems, and provide for the first time a real-world data-based test of one of the primary postulated  memory benefits of this universal spatial encoding scheme found in the mammalian brain.

The paper proceeds as follows. Section \ref{sec:background} provides an overview of data compression in signal processing with relevance to the approach presented here. Section \ref{sec:approach} describes the components of our proposed approach in detail. Experimental results and analysis are presented in section \ref{sec:experiment}, with Section \ref{sec:conclusions} discussing the findings and future areas of research.

\section{Background}
\label{sec:background}

Data compression has a broad range of applications in signal processing, which is to encode data into compact representations by taking advantage of perceptual and statistical properties of data to provide superior analytical results. In image processing, we can use cosine transform to compress a BitMap (BMP) image as a JPEG format with a tolerable information loss but a much smaller data size. In computer vision, images and videos are usually represented as high-dimensional visual feature vectors. The goal of encoding images into compact codes is to simultaneously reduce the storage cost and accelerate the computation. To achieve this, the most discriminant information contained in high-dimensional data is usually embedded into a lower-dimensional space for further analysis. Usually,  the embeddings are in a discrete format such as hashing \cite{TPAMI:HASH}. However, the discrete data representations suffer from data collisions when data size is large, so it is not the best option for unique mapping in visual place recognition. To avoid data collision, visual information can also be embedded in continuous, rather than discrete lower-dimensional spaces \cite{RAS:KDE}.

For multimedia data compression, there are two encoding families based on machine learning techniques: binary embedding and vector quantization, both of which are designed to compress continuous sensor data into discrete feature spaces. The idea of binary embedding is to represent feature vectors as compact binary codes, so the Euclidean distance between two vectors could be approximated by Hamming distance in the binary space \cite{TPAMI:HASH}. The advantage of binary embedding is due to the efficient Hamming distance computation, which can be implemented by the XOR and POPCOUNT operations. Different from binary embedding, vector quantization (VQ) adopts a codebook as a dictionary to quantize the feature vectors into a set of codewords, and the distances between any two codewords are pre-computed and stored in a lookup table \cite{BOOK:VQ}. When the original feature space is decomposed into the Cartesian product of several low-dimensional subspaces, vector quantization becomes product quantization (PQ) \cite{TPAMI:PQ, TPAMI:OPQ, TIP:BOPQ}. Compared with binary embedding, PQ based encoding methods have a lower information loss and thus can achieve a better accuracy, at the cost of a slightly lower computational speed. Both binary embedding and vector quantization are effective encoding techniques with regards to calculation and storage costs. Although binary embedding and vector quantization have different encoding strategies, the underlying mechanism is clustering, i.e., similar statistical patterns have the same codes. Both of the two techniques suffer from code collisions as the data size increases, and they have linear storage growth with the number of data instances.

Computational and storage requirements are of particular importance for mobile robotic and autonomous systems. It is important to differentiate between at least three different goals - achieving highly compact but ultimately linear storage growth, achieving sub-linear computational requirements (but with linear or worse storage growth), and achieving sub-linear storage growth. For the first goal, various techniques have been applied in robotic applications. For example, Locality Sensitive Hashing (LSH) is used to deal with the problem of stereo correspondence estimation \cite{ICRA15:LSH}, multi-model fusion techniques are adopted in humanoid robots to process large-scale language data \cite{ICRA12:BIGRAM}, and Local Difference Binary (LDB) descriptor is applied to obtain a robust global image description for place recognition and loop closure detection \cite{IROS2014:BINARY}. In \cite{ICRA16:TACTILE} the authors proposed to compress sensory data from tactile skins. Similarly, the distributed sensor data with high-frequencies can be compressed as coresets for streaming motion \cite{IPSN12:CORESET}. To achieve the sub-linear computation, \cite{ICRA17:SQLSLAM} builds an index on the original database to reduce the computation cost. While significant effort has been devoted towards efficient computation with low absolute storage costs, there has been little work examining how sub-linear storage growth might be achieved for SLAM systems, or examining how natural systems achieve this sub-linear storage growth with a unique one-to-many neuronal mapping system.

%Some methods such as \cite{ICRA17:SQLSLAM} solely build an index on the original database to reduce the computation cost but without data compression.    Some research outcomes in computer vision and natural language processing are also applied in robotic applications.  
%In \cite{ICWISE13:SENSOR} the authors introduced a local adaptive data compression based on Fuzzy transform to minimise bandwidth, memory space and energy consumed in radio communication. 

\section{Approach}
\label{sec:approach}

In this section, we describe our proposed encoding model for scalable place recognition, based on supervised learning techniques. The system comprises a periodic template learning phase, a database encoding phase, and a global place reconstruction phase.

\begin{figure}[t]
\centering
\includegraphics[width=0.45\textwidth]{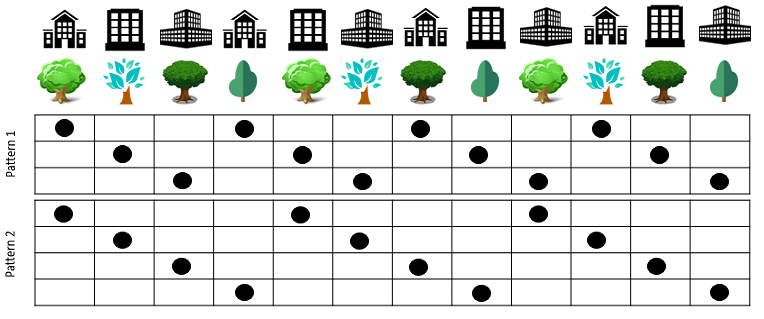}
\caption{An illustrative scenario with only two visual patterns available: buildings and trees, and both of them appear periodically. Each column represents a frame, so the frame sequence simulates a virtual camera moving forward. The combination of the two landmarks can represent at most $3\times 4=12$ distinct locations.}
\label{fig:fake_road}
\end{figure}

\subsection{Learning Periodic Patterns from Frame Sequences}

For clarity, we start with a toy example, since in the actual system we are looking for the underlying visual patterns that are often not intuitive. In Fig. \ref{fig:fake_road}, we show an illustrative scenario with only two visual patterns: different buildings and different trees. In this example, each column represents a frame in a sequence. By observing the frame sequence, we can see the style of building cycles in every 3 frames, and the type of the tree regularly changes in every 4 frames, respectively. Consequently, the combination of the two ideal periodic patterns can uniquely represent at most $3\times 4 = 12$ different locations. In place recognition systems, if there are more than two periodic patterns with different lengths, only storing the pattern information is sufficient for global place estimation. For example, we can use periodic landmarks and their positions to describe a frame, just like the visual templates used to detect interesting points as SIFT descriptors in image processing, and these descriptors can be then aggregated into visual feature vectors for further analysis. However, we will show that it is possible to learn latent periodic patterns from a wide variety of data. To enable the SLAM system to automatically analyse the feature vectors, in \cite{IROS17:DEJAVU} the authors proposed to apply spectrogram to find the regularly repeated patterns in frame sequences. The spectral methods assume the signal is composed of phase-shifted sine and cosine curves with scaled factors and offsets, but such a strong assumption is not valid in most real applications. Also, the thresholds need to be neatly set for signal discretization and template matching, because a high threshold may lead to the loss of matched templates, while a low threshold usually results in mismatches or redundancies.

In our proposed data compression model, temporally periodic patterns are learned from the frame sequence in a database. Given an (integer) period $\tau$, our system will look for a linear separation for each possible (integer) phase of that period that separates frames in that phase from frames in all other phases. This allows us to train completely distinct classifiers for each phase of a period, which is much less restrictive than training a single template for all phases of that period, as is implicit in spectral methods.

%In our proposed data compression model, temporally periodic patterns are learned from the frame sequence in a database. The advantage of learning-based methods over spectral ones is we do not need to assume cyclic properties are directly observable from the frame sequences. Instead, when given a frame sequence and a period length $\tau$, the model only needs to learn {\em how a frame can match a periodic template}. 

Let a location database be represented as a frame sequence as $\mathbf{X}=\{\mathbf{x}_1, \mathbf{x}_2, \ldots, \mathbf{x}_N\}$ where $N$ is the total number of data instances, and $\mathbf{x}_i$  is the $i$-th frame in the database ($1\le i \le N$). If each frame is represented as a $d$-dimensional visual vector, i.e., $\mathbf{x}_i\in\mathbb{R}^d$, the size of the database is $\mathbb{R}^{N\times d}$. When there is a cyclic visual pattern with the length $\tau$, $\tau$ templates are generated in each period. For the $j$-th template within a period ($1\le j \le \tau$), we assign a binary label $y_i^{(j)}\in\{1, -1\}$ for each frame $\mathbf{x}_i$ to indicate if it can match the $j$-th template: 
\begin{equation}\label{eq:lbl}\small
y_i^{(j)} = \begin{cases}
	1 & \quad i\mod\tau = j - 1, \\
	-1 & \quad \text{otherwise}. \\
\end{cases}
\end{equation}

Consequently, the task of determining whether a frame $\mathbf{x}_i$ can match the $j$-th template in a period becomes a binary classification problem. The weight vector $\mathbf{w}_j$ and bias $b_j$ can be simply obtained by solving a linear SVM as follows:
\begin{align}\small
\min_{\mathbf{w}_j,b_j,\xi_{ij}} \quad& \frac{1}{2}\|\mathbf{w}_j\|^2+C\sum^N_{i=1} \xi_{ij}, \nonumber \\
\mathbf{s.t.} \quad& y_i^{(j)}(\mathbf{w}_j^{\top}\mathbf{x}_i + b_j) \ge 1 - \xi_{ij}, \nonumber \\
	& \xi_{ij} \ge 0, 1\le i \le N, 
\end{align}
where $C$ is the penalty parameter to balance the hinge loss and functional margin. Here we mainly focus on the loss function to make the templates as linearly separable as possible, and we can simply set $C=\log{N}$. 

Simultaneously considering $\tau$ templates within a period, these binary classifiers can be integrated into a multi-class SVM model:
\begin{align}\small
\min_{\mathbf{w}_j,b_j,\xi_{ij}} \quad& \frac{1}{2}\sum_{j=1}^{\tau}\|\mathbf{w}_j\|^2+C\sum_{j=1}^{\tau}\sum^N_{i=1} \xi_{ij}, \nonumber \\
\mathbf{s.t.} \quad& y_i^{(j)}(\mathbf{w}_j^{\top}\mathbf{x}_i + b_j) \ge 1 - \xi_{ij}, \nonumber \\
	& \xi_{ij} \ge 0, 1\le i \le N, 1\le j \le \tau.
\end{align}

This multi-class SVM can be efficiently computed by some toolboxes such as scikit-learn\footnote{http://scikit-learn.org/}. The statistical property of the weight vector $\mathbf{w}_j$ is straightforward: when all sequenced frames within the database are represented by $\lfloor N/\tau \rfloor$ periods and can be perfectly segmented by $\tau$ classifiers, each classifier has the minimum covariance with its positively classified data instances. Thus, these classifiers can be just considered as the templates within a period.
%This multi-class SVM can be efficiently computed by some toolboxes such as scikit-learn\footnote{http://scikit-learn.org/} or LibLinear\footnote{https://www.csie.ntu.edu.tw/$\sim$cjlin/liblinear/}. The statistical property of the weight vector $\mathbf{w}_j$ is straightforward: when all sequenced frames within the database are represented by $\lfloor N/\tau \rfloor$ periods and can be perfectly segmented by $\tau$ classifiers, each classifier has the minimum covariance with its positively classified data instances. Thus, these classifiers can be just considered as the templates within a period.

The optimised weight vector $\mathbf{w}_j$ and bias $b_j$ ($1\le j \le \tau$) are able to determine if a frame is at the $j$-th position within a period with length $\tau$, i.e., the template that can be matched by a frame $\mathbf{x}$ is calculated by:
\begin{equation}\label{eq:index}\small
f (\mathbf{x} | \tau)=\arg \max_j (\mathbf{w}^{\top}_j \mathbf{x} + b_j), \quad 1\le j \le \tau.
\end{equation} 

Note that in order to keep the sub-linearity for data compression, we do not use kernel SVMs. In the kernel case, the weight vector in the decision function $f(\cdot)$ is represented as a linear combination of support vectors. Although the kernel decision function is more discriminative than the linear one, it cannot achieve the sub-linear data compression, because when either the dimension or the size of database increases, the number of support vectors also increases accordingly.

\subsection{Database Encoding}
\label{subsec:compression}

As we have seen, we can use linear SVMs to learn $\tau$ periodic templates given a period tau and the $N$ frames of a dataset in which we wish to localise. However, unless $\tau > N$, this is obviously not sufficient to uniquely identify a frame. The core idea of our method is to learn two or more such cyclic patterns $\{\tau_1, \tau_2, \ldots\}$, such that frames can be uniquely identified. In this subsection, we show how the periods $\tau$ can be chosen to allow for unique identification while minimising storage requirements. 

Assuming there are several candidates for $\tau$ available, we simply select $r$ cyclic patterns with periods $\tau_1,\tau_2,\ldots,\tau_r$ to estimate a frame position within a database when given an arbitrary frame $\mathbf{x}$ from it. The position of $\mathbf{x}$ could be determined by the {\em phase matches}, which are represented as a candidate set $\{j_1, j_2, \ldots, j_r\}$, $1\le j_k\le \tau_k$. The possible index $i$ is calculated by:

\begin{equation}\label{EQ:IDENT}
i=a_k\cdot\tau_k+j_k, 
\end{equation}
where $k\in\{1,\ldots,r\}$, and $a_k$ is a natural number. To identify $\mathbf{x}_i$ with $\{j_1, j_2, \ldots, j_r\}$, its index $i$ needs to be the unique solution of Eq. (\ref{EQ:IDENT}). Thus the selections of $\tau_1,\tau_2,\ldots,\tau_r$ should satisfy:

\begin{equation}\label{EQ:LCM}
lcm(\tau_1,\tau_2,\ldots,\tau_r)\ge N,
\end{equation}
where $lcm(\cdot)$ is the least common multiple operator. This condition guarantees the index mapping is unique, and there are sufficient ``slots" to store all frames in the database. 

\begin{figure}[h]
\centering
\includegraphics[width=0.25\textwidth]{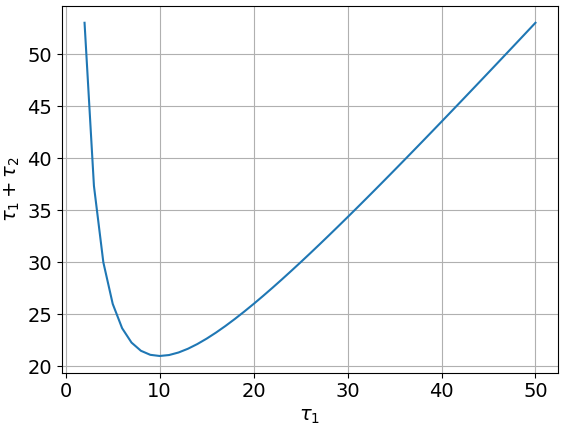}
\caption{The minimum number of templates required when $N=100$ and $r=2$.}
\label{fig:taus}
\end{figure}

If we manage to make $\tau_1,\tau_2,\ldots,\tau_r$ co-prime, then Eq. (\ref{EQ:LCM}) will be equivalent to $\prod\limits^{r}_{k=1}\tau_k\ge N$ . Given this constraint and if $r=2$, when given a $\tau_1$, $\tau_2$ needs to be at least $N/\tau_1$. In Fig. \ref{fig:taus}, we illustrate how the selection of the period lengths affects the total number of templates if $N=100$ and $r=2$. For each $\tau_1$ on the x-axis, we found the smallest $\tau_2$ that is both co-prime with $\tau_1$ and that satisfies $\tau_1\times\tau_2\ge N$. Then on the y-aixs we report $\tau_1+\tau_2$, which is proportional to the storage cost. We can see the minimum storage for 100 place estimations is 21 when $r=2$, so $\tau_1=10$ and $\tau_2=11$ would be the best period pair. 

%The x-axis represents the possible values of $\tau_1$, and the y-axis is the approximation of the total number of templates $\tau_1+\tau_2$, respectively.
%When we need to store $N$ locations with $r=2$, i.e., we only use two periodic patterns with lengths $\tau_1$ and $\tau_2$ for $N$ location estimations, the value of $\tau_2$ should be selected somewhere near $N/\tau_1$. 

Assuming that our place recognition algorithm could correctly identify the phase matches $j_k$ (see Eq. (\ref{EQ:IDENT})) corresponding to a query $\mathbf{x}$, the memory requirements for our system are to store the weight vectors $\mathbf{w}^{(k)}_1,\mathbf{w}^{(k)}_2,\ldots,\mathbf{w}^{(k)}_j$ and biases $b^{(k)}_1,b^{(k)}_2,\ldots,b^{(k)}_j$ for each possible $k$. In other words, we need to allocate memory for $\sum\limits_{k=1}^{r}\tau_k$ vectors of size $d+1$. Thus, the minimal storage requirements are achieved when $\prod\limits_{k=1}^{r} \tau_k \ge N$. In unconstrained cases, the solution is $\sqrt[r]{N}$, but $\tau_1,\tau_2,\ldots,\tau_r$ also need to satisfy they are coprime integers. Since we have not found a closed-form solution to this constrained problem, we instead propose to sample candidates from $\lceil\sqrt[r]{N}\rceil, \lceil\sqrt[r]{N}\rceil+1, \ldots,\lceil\sqrt[r]{N}\rceil+m$ with the least training errors. The training error $e$ is the fraction of misclassified training samples of $\tau$ linear models among the whole training set. Therefore, only storing $r$ groups of periodic templates can reduce the space complexity from $O(N)$ to $O(r\sqrt[r]{N})$.

%Thus, the minimal storage requirements are achieved if \tau_k=\arg\min \{\tau_1,\tau_2,\ldots,\tau_m\}$ under the condition  $\prod\limits_{k=1}^{r} \tau_k \ge N$, where $m$ is the number of sampled candidates for $\tau$. 

\subsection{Reconstructing a Global Place Estimate}
\label{subsec:search}

The localisation of an arbitrary frame from the database $\mathbf{x}$ is implemented by the intersection operation. We first calculate the phase matches of $r$ periodic patterns: $j_k=f(\mathbf{x}|\tau_k)$ for $k\in\{1,2,\ldots, r\}$ by applying Eq. (\ref{eq:index}) , and then generate $r$ candidate sets $P_1, P_2, \ldots, P_r$, where $P_k=\{f(\mathbf{x}|\tau_k), f(\mathbf{x}|\tau_k)+\tau_k, \ldots, f(\mathbf{x}|\tau_k)+\lfloor N/\tau_k \rfloor\}$. The index of $\mathbf{x}$ in the original frame database is calculated by $f(\mathbf{x}|\tau_1,\tau_2,\ldots,\tau_r)=\bigcap\limits_{l=1}^{r} P_k$.

%and Eq. (\ref{EQ:IDENT}) 

Before the online searching phase, given a query image, the system should first determine if the location that the image represents can be found in the database. Some appearance-based SLAM systems such as \cite{ICRA08:FABMAP}, proposed to set a lower bound of likelihood, which is calculated in the training procedure and set by users. The low likelihood that falls below the lower bound means the query image cannot match any places in the database. Our proposal is a deterministic approach, and there would be at least one phase match when given a query even if it is actually an outlier. So a lower bound of decision value can also be set to determine if a query can match a template in a periodic pattern. Alternatively, an auxiliary classifier could be trained when there are negative exemplars available, which are not descriptive to any locations in the database.

The retrieval of our method consists of $r-1$ 1d intersections of $r$ sets with size $N/\tau$, which is lower than $\sqrt[r]{N}$. Since the sets are already sorted, the 1d intersection can be achieved in $O(r\sqrt[r]{N})$: the intersection of a pair of sorted sets is linear in the sum of the sizes of both sets, so its time complexity is $O(\sqrt[r]{N})$, and we only need to perform $r-1$ such intersections on all sets.

\subsection{Improving Robustness Through Multi-Exemplar Training and Augmentation}
\label{subsec:data_env}  

A common method for improving the robustness of recognition techniques is to use, where available, multiple different examples of an object or place in training. For example, ImageNet has over ten million images with one thousand categories \cite{IJCV:IMGNET}. With large-scale and well-labelled training images, the classifiers trained on such datasets have near-human performance in very challenging recognition tasks. For the mobile place localisation systems, the robot should be able to memorise the scenes by revisiting the same places multiple times from different perspectives, or under distinct appearance conditions, to improve their discriminative power. In this case, the periodic patterns are essentially learned in a shared space rather than the original feature space.

Since our proposed data compression model for scalable place recognition is also based on supervised learning techniques, using frames taken under different appearance conditions has the potential to improve recognition accuracy and robustness. However, there is not always multi-exemplar training data available. In this case, we can apply image augmentation methods such as Gaussian blur, flipping, random cropping and elastic transformation to simulate the multi-exemplar data environment.

\section{Experiment and analysis}
\label{sec:experiment}

In this section, we describe the datasets used, the image preprocessing methods, the evaluation metrics and the experimental results. We also provide analysis of our proposed model, breaking down the performance contributions of the core system, enhancements including multi-exemplar training and augmentation, and provide an analysis of the trade-off between performance and storage scaling.

\subsection{Datasets and Experiment Settings}

To evaluate our proposed data compression model for scalable visual place recognition, we experimented with three different datasets: Nordland Train dataset, Aerial Brisbane dataset and Oxford RobotCar dataset.

\subsubsection{Nordland Train dataset}

The Nordland Line\footnote{https://nrkbeta.no/2013/01/15/nordlandsbanen-minute-by-minute-season-by-season/} is a 729-kilometre railway between Trondheim and Bod\o, Norway. This dataset contains four long videos captured by placing a camera at the front of a train facing forward along the railway track. The four videos describe the front views in four seasons, and each video is about ten hours long. This dataset was first used for visual navigation across seasons\cite{ECMR13:ACLCS}, but we only use it to test the compressibility of our proposed model. In the first part of our evaluation, the queries are all from the reference data, and the model aims to find the exact positions of them in the database. To pre-process the video data, we first extracted the keyframes, then used the optical flow of the ground directly in front of the train to estimate the velocity then normalised it. Finally, the four subsets contain 10,713, 7,403, 9,267 and 7,276 frames, respectively.

\subsubsection{Aerial Brisbane dataset}
The Aerial Brisbane dataset is generated by taking a snapshot from NearMap\footnote{http://maps.au.nearmap.com/}, which describes the Brisbane region in Queensland, Australia. The total size of the image is $7526\times6562$, and each pixel is an actual geographic area of $4.777\times 4.777$ square metres. The image was then segmented to $224\times 224$-pixel frames with 112-pixel strides, so in our setting the dataset can represent 3,705 different places. 

We use this dataset to test if our model can recognise locations in visual changing environment. To simulate the environment, the query images and reference data are from different sources. We collected several snapshots of the aerial map taken at different times ranging from 19/05/2013 to 24/06/2017. One image was selected as the query to search the absolute locations of the patches on the map, and the rest reference frames are used to train the model.

\subsubsection{Oxford RobotCar dataset}
The Oxford RobotCar dataset \cite{IJRR:RCD} contains over 100 repetitions of a consistent route through Oxford, UK. The dataset captures many different combinations of weather, traffic and pedestrians. For our testing, we used 5 subsets of the Oxford RobotCar dataset generated from a fixed route captured at different times of day using images captured by the Point Grey Bumblebee XB3. This dataset describes a small area with a limited number of locations. Since the geographic positions are described in consecutive northing and easting values as GPS data, we applied KMeans on the normalised coordinates to generate 100 clusters, so the GPS coordinates falling into the same cluster are considered as a unique location on the map. We ran our place recognition model to test if the periodic encoding can successfully capture the cyclic properties of the frame sequences to enable accurate reconstruct the location estimation.

For all of the three datasets, we utilised the deep visual features extracted from popular ConvNet architecture to describe the frames. Specifically, we used the second output of the fully-connected layer of the VGG16 model \cite{ARXIV:VGG} and then applied L2 normalisation. Thus each frame is represented by a 4,096d visual feature vector. On the Oxford RobotCar dataset, each location is visually represented by multiple frames. To reduce the noise and fit class conditional densities to the data with multiple exemplars, we further applied the Linear Discriminant Analysis (LDA) to reduce the dimensionality to 64.

In our experiment, we compared our model with KD-Tree\cite{COMM75:KD_TREE}, Iterative Quantization (ITQ)\cite{TPAMI:ITQ} and Optimised Product Quantization (OPQ)\cite{TPAMI:OPQ} on Nordland Train and Aerial Brisbane datasets. KD-Tree is a well-known method for approximate nearest neighbour search, which builds a binary search tree as the index for a fixed-sized database. Although a KD-Tree can effectively accelerate the computation, this technique does not compress the data. ITQ and OPQ are discrete embedding approaches, which encode the high-dimensional data into compact codes for fast computing. However, they cannot achieve unique mapping for place recognition because code collision is inevitable. Furthermore, these techniques compress data in an absolute way, i.e., the compressed data size is proportional to the actual data size, which is not sub-linear. 

Let $T$ be the level of the period length, which is the minimum value for $\tau$ in data encoding. Assume we set $T=2\sqrt{N}$ and $r=2$ for our model, and use a 256-bit binary vector and a 256d integer vector to encode a 4,096d visual instance for ITQ and OPQ, respectively. When data size is comparably small, our proposed model has a higher memory cost, but it increases sub-linearly when the database becomes extremely large.

%When the size of database increases linearly, the curve of memory usage is displayed in Fig. \ref{fig:mem_curve}.  

%\begin{figure}[h]
%\centering
%\includegraphics[width=0.25\textwidth]{fig/mem_curve.png}
%\caption{The theoretical memory cost comparison of different data compression models.}
%\label{fig:mem_curve}
%\end{figure}

We used the compression ratio to demonstrate how our model can achieve sub-linear storage and used the accuracy metric to evaluate place recognition performance. We also compared the computational speed of our model with the baseline models. 

Our experiment was conducted on a desktop with Intel(R) i7-7700K CPU \@4.20GHz with 4 processors, 32GB RAM, and Windows 10 operating system with a Python 3.6 computational environment.

\subsection{Place Recognition Results When All Queries Are from Reference Data}
\begin{figure}[t]
\centering
\includegraphics[width=0.25\textwidth]{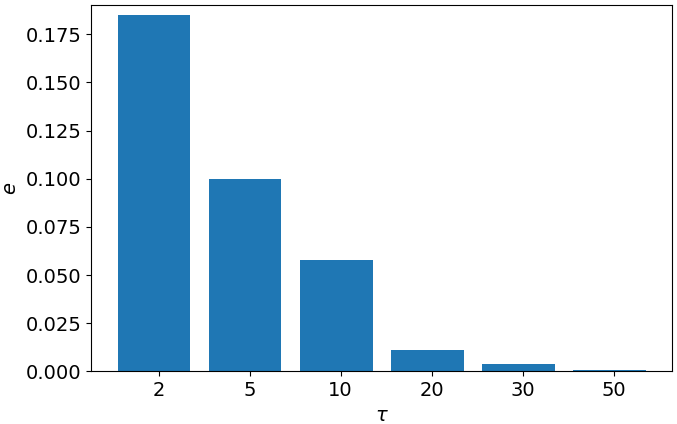}
\caption{The training errors on Aerial Brisbane dataset when setting different period lengths.}
\label{fig:period_err}
\end{figure}

%To encode a fixed database, we chose the period lengths $\tau_1,\tau_2,\ldots,\tau_r$ to control the compression rate.
We first investigated how the period length $\tau$ affects the training error. We tested different lengths of periods and trained the linear SVMs on the Aerial Brisbane dataset, with the training errors illustrated in Fig. \ref{fig:period_err}. We can see longer periods lead to a lower training error rate $e$ and a higher compression ratio when $r$ is fixed. In the extreme case, when a whole frame database has only one period, i.e., $r=1$ and $\tau=N$, no data compression is implemented, and the model reverts to brute-force search.

We then investigate the data compression results for Nordland Train and Aerial Brisbane datasets when only two periodic patterns are available, i.e., $r=2$. We also tested the training errors at two periodic levels: $T=\sqrt{N}$ and $T=2\sqrt{N}$, respectively. For each subset of Nordland Train dataset, as well as the Aerial Brisbane dataset, we tried seven different values of $\tau$ in training the linear SVMs and recorded the error rates, and then the system automatically selected the best period pair. Based on the selection of periods, we analyse the storage cost when applying our proposed data compression approach and selecting the length of period $T$ at the level of $\sqrt{N}$ on the two datasets. In a 32-bit operating system, the memory cost for a float number is 4 bytes. The storage comparison of the datasets is summarised in Table \ref{tab:data_compression}. From the table, we can see that our proposed model is able to encode very large frame databases with high compression ratios. If the data size increases linearly, applying two-period values and several templates can make the storage increase in a sub-linear manner. When the number of frames is more than 10,000, our model only takes about 1/50 memory to store all data instances.

\begin{table*}[t]
	\centering
	\caption{The data compression results ($r=2$ and $T=\sqrt{N}$). Note the 4097 in the Compressed size column comes from 4096 plus the bias.}
		\label{tab:data_compression}
		\begin{tabular}{|c||c|c|c|c|c|}
		\hline
		{\bf Dataset}	&{\bf Original size} &{\bf Original storage} &{\bf Compressed size} &{\bf Compressed storage} &{\bf Compression ratio}\\
		
		\hline
		Norland (spring) &$\mathbb{R}^{10713\times 4096}$ &175,521,792 bytes &$\mathbb{R}^{211\times 4097}$ &3,457,868 bytes &0.0207  \\
		\hline
		Norland (summer) &$\mathbb{R}^{7403\times 4096}$ &121,290,752 bytes &$\mathbb{R}^{181\times 4097}$ &2,966,228 bytes &0.0246 \\
		\hline
		Nordland (fall) &$\mathbb{R}^{9267\times 4096}$ &151,830,528 bytes &$\mathbb{R}^{201\times 4097}$ &3,293,988 bytes &0.0217\\
		\hline
		Nordland (winter) &$\mathbb{R}^{7276\times 4096}$ &119,209,984 bytes &$\mathbb{R}^{179\times 4097}$ &2,933,452 bytes &0.0246  \\
		\hline
		Aerial Brisbane &$\mathbb{R}^{3705\times 4096}$ & 60,702,720 bytes &$\mathbb{R}^{125\times 4097}$ &2,048,500 bytes &0.0337  \\
		\hline
		\end{tabular}	
\end{table*}

We used the frames from the reference data as queries and applied our models for location estimation. We show the place recognition results of Nordland Train and Aerial Brisbane datasets in Table \ref{tab:acc_nordland_aerial}. In all of the four subsets of the Nordland Train dataset, none of the accuracies falls below 98\% even when we apply the extreme compression method. For example, in the spring subset, 10,713 frames record diﬀerent views of places along the Norland railway. Applying our proposed data compression model can still achieve 99.46\% accuracy. If we set longer periods, i.e., $T=2\sqrt{N}$, the compression ratio doubles, but the recognition accuracy is higher, which is very close to 100\%. On the Aerial Brisbane dataset, our system achieved the very near-perfect accuracy of 99.92\% (only 3 mismatches) when $T=\sqrt{N}$. When the length of period doubles, the recognition accuracy is 100\%. 

\begin{table*}[t]
	\centering
	\caption{Best learned periods and place recognition results on Aerial Brisbane dataset and Nordland train dataset ($r=2$ and $T=\sqrt{N}$). Note all queries are from the reference data.}
		\label{tab:acc_nordland_aerial}
		\begin{tabular}{|c||c|c|c|c|c||c|c|c|c|c|}
		\hline
		\multirow{2}{*}{{\bf Dataset}}	& \multicolumn{5}{|c||}{{\bf $T=\sqrt{N}$}} & \multicolumn{5}{|c|}{{\bf{$T=2\sqrt{N}$}}}\\
		\cline{2-11}
		&$\tau_1$ 	&$e$ ($\times 10^{-3}$)	&$\tau_2$ 	&$e$ ($\times 10^{-3}$)	&Accuracy	&$\tau_1$ 	&$e$ ($\times 10^{-3}$)	&$\tau_2$ 	&$e$ ($\times 10^{-3}$)	&Accuracy \\
		\hline
		Norland (spring)&105	&2.99	&106	&2.89	&0.9946	&211	&0.84	&212	&0.65	&0.9990	\\
		\hline
		Norland (summer)&88	&8.91	&93	&8.65	&0.9846	&179	&2.70	&180	&2.30	&0.9960	\\
		\hline
		Norland (fall)&98	&4.64	&103	&4.64	&0.9912	&197	&0.54	&200	&0.22	&0.9992	\\
		\hline
		Norland (winter)&87	&1.51	&92	&1.52	&0.9971	&174	&0.27	&175	&0.00	&0.9996	\\
		\hline
		Aerial Brisbane&62	&0.54	&63	&0.27	&0.9992	&122	&0.00	&123	&0.00	&1.0000	\\
		\hline
		\end{tabular}	
\end{table*}

\begin{figure}[t]
\centering
\subfigure[Compression ratios]{
\begin{minipage}[b]{0.21\textwidth}
\includegraphics[width=1\textwidth]{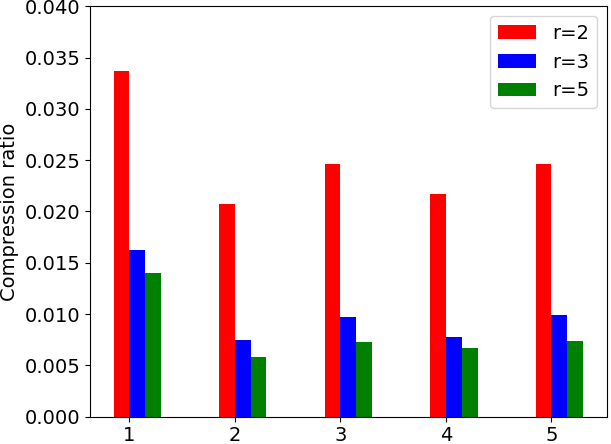}
\end{minipage}
}
\subfigure[Accuracy comparisons]{
\begin{minipage}[b]{0.21\textwidth}
\includegraphics[width=1\textwidth]{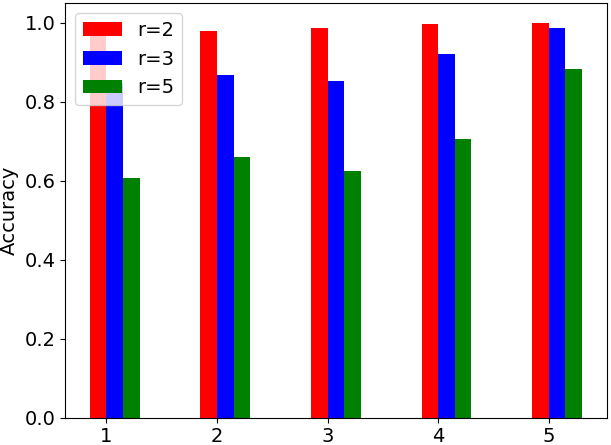}
\end{minipage}
}
\caption{The compression ratios and accuracy comparisons when $r$ changes. Note the queries are all from reference data. (1. Nordland spring; 2. Nordland summer; 3. Nordland fall; 4. Nordland winter; 5. Aerial Brisbane)}
\label{fig:compressions}
\end{figure}

We conducted an experiment using both Nordland Train and Aerial Brisbane datasets evaluating the performance of the KD-Tree, ITQ, OPQ, and the brute-force search techniques, recording the average search time. The comparison is displayed in Table \ref{tab:time}. By using a few learned periodic templates and the matching approach introduced in section \ref{subsec:search}, the computational efficiency is significantly increased compared to the exhaustive search, although it is a lower than KD-Tree, ITQ and OPQ. However, KD-Tree only builds an index on the database but does not implement the data compression at all. Both ITQ and OPQ could compress the data in an absolute manner, but they cannot achieve the unique mapping required for place recognition, even when the distance between a query and its matched frame is zero. Considering the compression ratio, search accuracy and speed, it is worth applying our proposed model to large-scale place recognition systems.

\begin{table}[h]
	\centering
		\caption{The comparisons of different search methods (the seach time is recorded on Aerial Brisbane dataset). }
		\label{tab:time}
		\begin{tabular}{|c||c|c|c|}
		\hline
		{\bf Method}	&{\bf Compressed scale} &{\bf Search time} &{\bf Unique mapping} \\
		\hline
		KD-Tree &No compression &0.000215 &Yes   \\
		\hline
		ITQ &Linear &0.000116 &No  \\
		\hline
		OPQ &Linear &0.000241 &No   \\
		\hline
		Brute-force &No compression &0.125948 &Yes  \\
		\hline
		Ours &Sub-linear &0.000503 &Yes   \\
		\hline
		\end{tabular}
\end{table}

Then we used more than two periods by setting $r=3$ and $r=5$ respectively and re-ran our model on Nordland Train and Aerial Brisbane datasets. As is discussed in section \ref{subsec:compression}, we can choose different available periods for data compression. A larger value of $r$ means the system can achieve a lower compression ratio. Fig. \ref{fig:compressions} summarises the compression ratios and the accuracy comparisons. From the two figures, it can be seen that although applying more periodic patterns can achieve an even higher compression ratio, the recognition accuracy falls significantly, even though the queries are all from reference data. The reason is that the value of $\tau$ is in direct proportion to the number of negative data instances in Eq. (\ref{eq:lbl}), i.e. when the number of templates increases within a period, there are fewer positive-labelled instances in the learning process, which makes the periodic templates more linearly separable. By contrast, the training error rate is higher on a more ``balanced'' dataset.

\begin{figure}[h]
\centering
\subfigure[Aerial Brisbane]{
\begin{minipage}[b]{0.22\textwidth}
\includegraphics[width=1\textwidth]{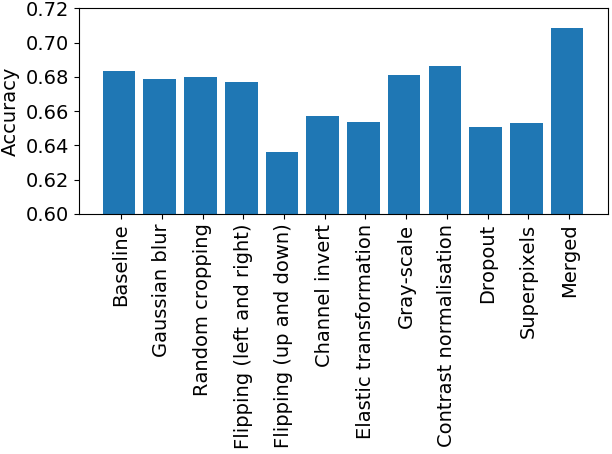}
\end{minipage}
}
\subfigure[Oxford RobotCar]{
\begin{minipage}[b]{0.22\textwidth}
\includegraphics[width=1\textwidth]{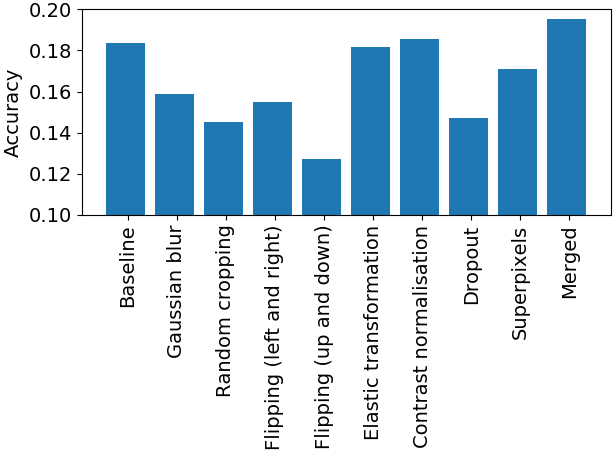}
\end{minipage}
}
\caption{The accuracies of different data augmentations.}
\label{fig:data_aug}
\end{figure}

We tested our system using data from different times of day for the Aerial Brisbane and Oxford Robotcar datasets. For Aerial Brisbane dataset, we used one image from this dataset as queries, and used another image taken at a different time as the reference. For Oxford RobotCar dataset, we used the frames taken on a distinct date as queries to search their locations. When applying the brute-force search, the accuracy is 0.9582 on Aerial Brisbane dataset, and 0.236 on Oxford RobotCar dataset, respectively. Using our model to compress the reference data, the accuracy dropped to 0.6835 and 0.183 on the two datasets. The result signifies solely applying the data compression model cannot achieve a satisfactory recognition accuracy under different appearance conditions. As is introduced in Section \ref{subsec:data_env}, we adopted different data augmentation approaches, including Gaussian blur, random cropping, flipping, elastic transformation, contrast normalisation, etc. We separately experimented with these augmented data sources, then merged them as a whole training set to train a unified place recognition model. Note that for Oxford RobotCar dataset, we did not use channel invert and grey-scale augmentations because the raw image data is grey-scale. The accuracies are displayed in Fig. \ref{fig:data_aug} for the two datasets. It can be seen that although each single augmented data source has limited power to help obtain a discriminative model, their combination can effectively boost the accuracy by 2\% and 1\% on the two datasets, respectively.

\subsection{Results in Visually Changing Environments}

\begin{figure}[h]
\centering
\subfigure[Aerial Brisbane]{
\begin{minipage}[b]{0.22\textwidth}
\includegraphics[width=1\textwidth]{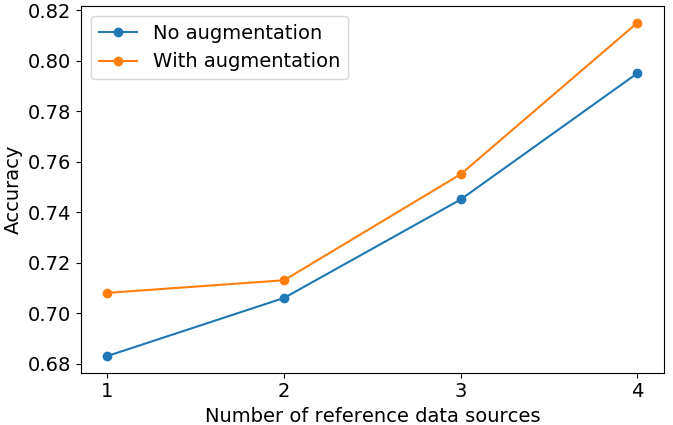}
\end{minipage}
}
\subfigure[Oxford RobotCar]{
\begin{minipage}[b]{0.225\textwidth}
\includegraphics[width=1\textwidth]{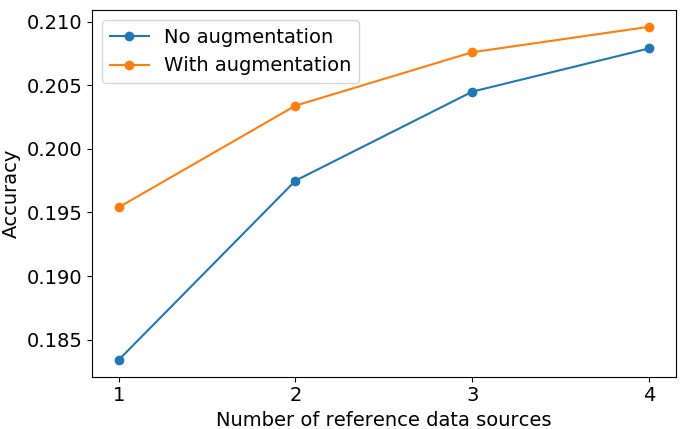}
\end{minipage}
}
\caption{The accuracy curves when there are multiple reference data sources ($r=2$).}
\label{fig:multi_src}
\end{figure}

Next, we trained the models with the multi-reference set, where these frames are taken under different appearance conditions. We used different combinations of reference sources and evaluated their performance on Aerial Brisbane and Oxford RobotCar datasets, and the accuracy curves are plotted in Fig. \ref{fig:multi_src}. The accuracy improves steadily as the number of data sources increases and learning with the multi-source data and image augmentation enables the data compression model to better deal with the various appearance conditions while keeping the storage sub-linear.

\begin{figure}[h]
\centering
\subfigure[Aerial Brisbane]{
\begin{minipage}[b]{0.22\textwidth}
\includegraphics[width=1\textwidth]{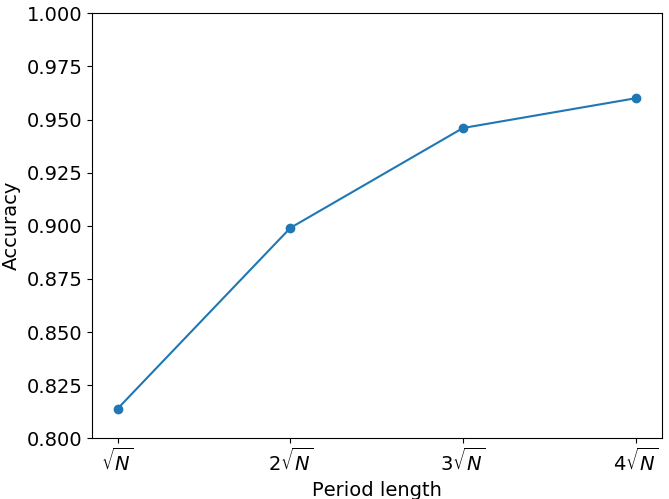}
\end{minipage}
}
\subfigure[Oxford RobotCar]{
\begin{minipage}[b]{0.22\textwidth}
\includegraphics[width=1\textwidth]{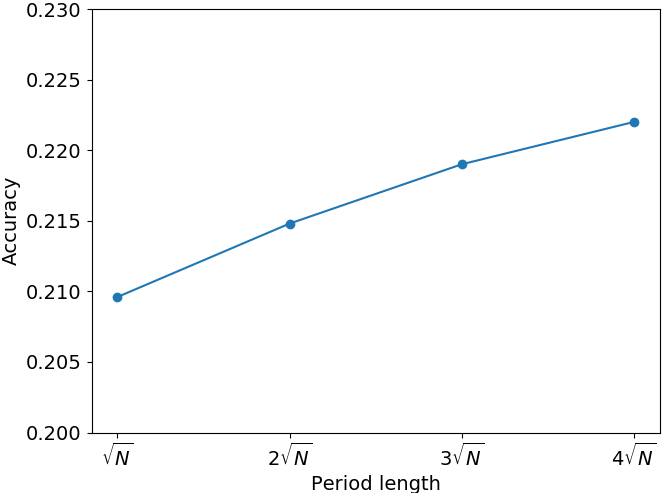}
\end{minipage}
}
\caption{The accuracy curve when setting different lengths of periods ($r=2$, 4 data sources with augmentation).}
\label{fig:multi_period_len}
\end{figure}

As is discussed in Section \ref{subsec:compression}, we set longer periods to further reduce the training error $e$, but with a lower data compression ratio. We tested different lengths of periods by setting $T=\sqrt{N}$, $T=2\sqrt{N}$, $T=3\sqrt{N}$, and $T=4\sqrt{N}$ respectively, and the accuracy curve is plotted in Fig. \ref{fig:multi_period_len}. We can conclude that setting a longer period can improve the recognition accuracy, with the sacrifice of the compression rate. In this real application scenario, if we set $T=4\sqrt{N}$, the data compression ratio is 0.131 on Aerial Brisbane dataset and 0.81 on Oxford RobotCar dataset, respectively, but the recognition performance is boosted.

\section{Discussions and conclusions}
\label{sec:conclusions}

We have presented a novel image-based map encoding scheme that deliberately seeks out and learns mutually supportive visual pattern frequencies in the environment to enable place recognition with sub-linear storage growth as the environment size increases. The system is based on the nature of neural mapping systems in the mammalian brain that does not appear to approach the data association problem central to most robotic mapping systems the same way; instead each neural map ``unit'' is associated with an arbitrarily large number of places in the environment distributed at regular intervals. 

Results on large real-world datasets show that the fundamental premise is valid and that high-performance place recognition can be achieved with a mapping system whose map storage scales sub-linearly with environment size. The system is agnostic of any particular types of features or feature frequencies and its performance across a range of environments shows that, perhaps surprisingly, repetitive visual patterns can usually be found. 

We applied the data augmentation and multi-source training data, which are generic methods for visual recognition tasks, to make our model more applicable under different appearance conditions. In future work, we could also design a more sophisticated system by integrating some advanced machine learning techniques to better capture the spatial properties of the periodic patterns and improve the recognition performance. Alternatively, we could apply some existing matching schemes such as SeqSLAM \cite{ICRA12:SEQSLAM}, to further improve the stability by taking consideration of multi-frame integration. 

\section*{Acknowledgements}
\label{sec:acknowledgements}

This work was supported by an Asian Office of Aerospace Research and Development Grant FA2386-16-1-4027 and an ARC Future Fellowship FT140101229 to MM.

\bibliographystyle{IEEEtran}
\bibliography{reference}

\end{document}